# Indonesia's Fake News Detection using Transformer Network

Aisyah Awalina

Brawijaya University, Computer Science, aisyahawalina13@student.ub.ac.id

Jibran Fawaid

Brawijaya University, Computer Science, jibranfawaid@student.ub.ac.id

Rifky Yunus Krisnabayu

Brawijaya University, Computer Science, rifkysentrix@gmail.com

Novanto Yudistira

Brawijaya University, Computer Science, yudistira@ub.ac.id

Fake news is a problem faced by society in this era. It is not rare for fake news to cause provocation and problem for the people. Indonesia, as a country with the 4th largest population, has a problem in dealing with fake news. More than 30% of rural and urban population are deceived by this fake news problem. As we have been studying, there is only few literatures on preventing the spread of fake news in Bahasa Indonesia. So, this research is conducted to prevent these problems. The dataset used in this research was obtained from a news portal that identifies fake news, turnbackhoax.id. Using Web Scrapping on this page, we got 1116 data consisting of valid news and fake news. The dataset can be accessed at https://github.com/JibranFawaid/turnbackhoax-dataset. This dataset will be combined with other available datasets. The methods used are CNN, BiLSTM, Hybrid CNN-BiLSTM, and Bert with Transformer Network. This research shows that the Bert method with Transformer Network has the best results with an accuracy of up to 90%.

**ACM Reference Format:**

# 1 INTRODUCTION

In the era of globalization, the internet has been used in the world widely. People can get a lot of information through various site. This information can be obtained through news portals, articles on the internet, and social media. Unfortunately, not all of these news and articles contain correct and valid information. This invalid information is called "fake news". Fake news is false information that is formed like a format of news media content without proper processing by the news agency [1].The spread of fake news is an important concern. Fake news could be a threat to journalism and freedom of expression. The fake news can be a provocation or slander against an important organization or person. A study from Stanford University discussed social media and fake news during the US presidency in 2016. This study revealed that fake news sites had been visited up to 159 million times during that time [2].

There is a wide variety of literature that develops ways of preventing the spread of fake news. These methods are text classification, network analysis, and Human-Machine Hybrid integration. As the implementation had been through, these methods have their own advantages and disadvantages in their implementation [3].Indonesia is one of the countries that facing the widespread of fake news. A survey showed 46% to 61% of rural knew provided fake news. Uniquely, the urban has a higher proportion than the rural. And there are 38% to 61.1% of rural believe the fake news and 45.3% to 79.6% of the urban believe the fake news [4]. This is sufficient to prove that the spread of fake news in Indonesia is quite concerning. This study will discuss one of the ways that can be used to indicate fake news. Using the Bahasa Indonesia as an instrument, we use several methods to detect fake news.

With such a large spread, a system that can detect thousands or perhaps millions of data at a time is needed. The amount of data scale, distribution, diversity, and / or timeliness require, shows that this is a field of Big Data Analytic [5]. Any technology that can be applied properly can have an advantage. Unfortunately, big data is very difficult to deal with [6]. So that in applying it, we need the right method in analyzing it.We compiled various methods that can be used in deep learning to indicate fake news. Study A. Wani, *et. al*. [7] discusses the detection of covid-19 fake news. Using the dataset Constraint @ AAAI2021 - COVID19 Fake News, the dataset contains news in English. The method used is two approaches, namely the simple model approach of deep learning (CNN, LSTM, BiLSTM + Attention, HAN) and the Transformer based model (BERT and DistilBERT). This study results in Transformer based having a better level of accuracy. Despite the results of other methods, CNN doesn't have poor accuracy either, it was 93.5%.

There are only few systems about detecting fake news in other languages than English. Study conducted by Hoossain, et. al., using a Bangladesh language dataset with 3 approaches, namely baseline, traditional linguistic features with SVM, and neural network models (CNN, LSTM, and BERT) [8]. This study produces a neural network model with CNN that has good accuracy with a total news class with a precision of 98%, a recall of 10% and an F1-Score of 99%.Another study that discusses the detection of fake news is a study provided by Jiang, T., *et. al*. [9]. This study uses a dataset consisting of 44898 articles with 21417 original news stories and 23481 fake news stories with two classes in it. Using the BiLSTM method and text embedding Glove, the accuracy obtained is 99.82%.Based on these studies, we used the BiLSTM and transformer as the hope of having better accuracy. This study also focused on detecting fake news in Bahasa. In addition, we use datasets that available on the internet and also use datasets that obtained through web scrapping on the turnbackhoax.id website.

## 2 RELATED WORK

The first research was conducted by Kaliyar in 2018 [10] focused on fake news on social media. The dataset used in research such as from Kaggle. In this study, machine learning algorithms such as naïve Bayes, decision trees and deep learning are used, such as shallow convolutional neural network (CNN). In this study also used feature extraction such as TF-IDF and then also used word embedding. The results obtained using CNN yielded an accuracy of up to 98.3%.

The next research on fake news was carried out by Ahn [11]. The data set was collected via the internet with a total of 50000 million with a 50:50 division between classes. In the signing process, the word piece model (WPM) method is used. The model used in this study is BERT as one of the state-of-the-art models. The results obtained in this study resulted in an ROC curve value of 0.838. So that the model built was good at detecting fake news.Other research that we have been read is a the detection of fake news using a semi-supervised learning approach [12]. In this study, using a dataset with a total of 50 rows of data with 75 data with fake labels, and 75 with real labels. Then for embedding in the research using GloVe pre-trained. Then the algorithm model used is Graph Convolutional Networks and Attention Graph Neural Network. The results show that many of these algorithms have better results when compared to machine learning algorithms such as SVM and Random Forest. The highest accuracy is obtained with the number 84.94%.

Research related to fake news was carried out by Ahuja [13]. In this study, using the dataset available in Kaggle. The total dataset is 12953 for fake news and 15712 for news labeled true / real. In this study using the GloVe embedding. Then the proposed algorithm uses a Hierarchical Attention Network added with stacked bidirectional Gated Recurrent Units (GRU). The results show that the proposed model produces a better model than other deep learning models such as LSTM, CNN, Bi-LSTM. The proposed model succeeded in obtaining an accuracy of 93.83% while the comparison algorithm, for example LSTM, only obtained 90.31%.

The next research was carried out by Aslam [14]. In this research, using the "LIAR" dataset in which there are 12.8K, around 2053 for data labeled true and 2504 for fake labels were selected so that the total dataset reached 4557. In the preprocessing stage using wordnet as tokenizer and WordNetLemmatizer as lemmatization. The algorithm used is a deep learning ensemble model, namely Bi-LSTM-GRU and dense deep learning. The result is an accuracy of 89.8%. Which shows better results than other algorithms such as LSTM and CNN.

The next research related to fake news was carried out by Girgis [15]. In this study, the WILD dataset contains 12,836 statements. The models used to detect fake news are recurrent neural networks (vanilla, GRU) and long-short-term memories (LSTM). The model built produces an accuracy that is around 21.7%.The next research was conducted by Fang in 2019 [16]. The dataset used in this study comes from fakenews.mit.edu where there are 12,228 fake news and 9,762 non-fake news. In this study, using word embedding, GloVe and Word2vec. Then this study uses a model of self-multi-head attention and convolutional neural network (SMHA-CNN). The model was evaluated using a 5-fold CV with an accuracy of being able to reach 95.9%.

The research we conducted is using Bahasa Indonesia as the instrument. A research that related to similar case is the research from [17]. The research is case folding, tokenizing, stopword removal, and term frequency at the preprocessing stage. After that, this research uses Naïve Bayes in its main process. Using 600 news items, the average accuracy obtained is 82.6 %. Dynamic testing was carried out with 60 news and resulted in a conformity of 68.3%.

Another research that uses Bahasa Indonesia as a dataset is research from [18]. This research uses a method that similar to the previously described research, they are Reader feedback and Naïve Bayes method. The author of this study considers Naïve Bayes with 10-fold cross validation to be the method that has the best accuracy based on the literature collected. Thus, by using these two methods and the dataset obtained through https://turnbackhoax.id with 250 news, this research produces an accuracy of 87%, 91% of precision, and 95% of f-measure. According to the lack of research on fake news detection in Bahasa Indonesia and the lack of methods used. This study offers a different method from the other fake news detection research in Bahasa Indonesia. We also use transformer on this research. We use it because based in literature, it has an effect on increasing the accuracy. In addition, this research conducted a scrapping on the https://turnbackhoax.id website with more news than other research to get a better tough result.

## 3 METHODOLOGY

### 3.1 BERT

BERT or Bidirectional Encoder Representations from Transformers is a form of model in transformers that applies bidirectional training then combines the context from the left and right layers [19]. RNN and Transformer have a similar architecture in that there are encoders and decoders. BERT applies a transformer architecture, where the focus is Attention. Attention is a focus where we determine the main focus or context of a sequence. For example, instead of just translating between languages, the encoder will also write down important keywords/words. Attention is also a function used to map queries and follow key-value and output pairs, all of which are in vector form [20]. Equation of Attention can be seen in Eq (1).

$$Attention(Q,K,V) = softmax\left(\frac{QK^T}{\sqrt{d_k}}\right)V \quad (1)$$

Where $Q$ is the matrix that builds the query (contains a vector of each word), $K$ is the key, $V$ is the value itself. *Softmax* is one of the activations in deep learning which is useful for the weight of attention leading to between 0 and 1 (a kind of probability).

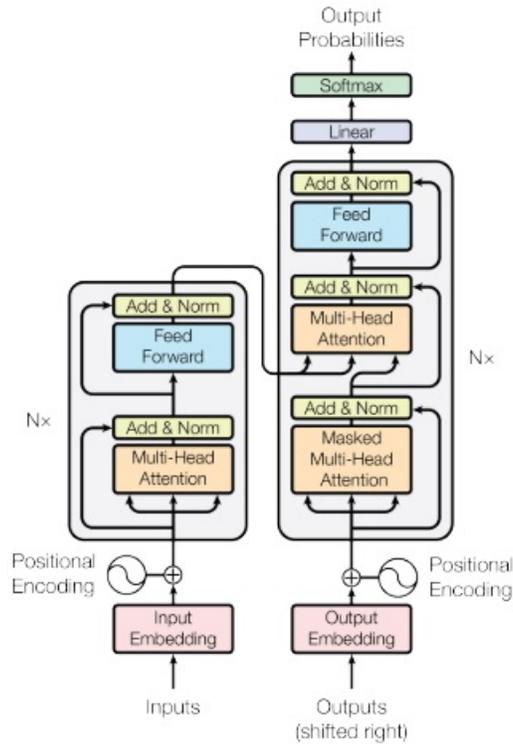

Figure 1: Architecture of Bert [20]

Basically, BERT can also be understood as a stack of transformer architecture. BERT has two models, BERT Base which has 12 layers of encoder layer stacks and BERT Large which has 24 stacks of encoder layers. BERT Base has about 110M parameters while BERT Large has 340M parameters[19]. In BERT there are two steps that are followed, called pre-training and fine-tuning. Pre-training is the stage where the model is trained using unlabeled data (unsupervised task) through pre-training tasks, while fine-tuning is a model initialized with the parameters of previous training results, and fine-tuning using data that is already labeled. In the pre-training, there is one task, namely the LM Mask where 15% of the words will be changed by a [MASK], then the model will make predictions in the form of a probability value for the possible words that can be filled. The second task, namely Next Sentence Prediction (NSP), is a mechanism for predicting the next sentence from a previous sentence. Where during the training process 50% percent of the input came from the original document, while 50% was chosen randomly. The purpose of training using these two tasks is to produce the smallest loss function value.

Table1: Hyperparameter for BERT

| Hyperparameter | Value |
| --- | --- |
| Loss function | Categorical-crossentropy |
| Learning rate | 2e-05 |
| Optimizer | Adam |

| Hyperparameter | Value |
|---|---|
| Number of epochs | 50 |
| Batch size | 16 |

## 3.2 CNN

Convolutional neural network (CNN) is a deep learning method that can be applied to classify text documents. The first layer is the embedding layer where the input results come from the word2vec process. We use 1 layer of Conv1D layer with kernel size 2. Conv1D is used because it can Convd1 can be used to process natural language. The Detail of CNN architecture that we use in this study can be seen in the Table 2.

Table2: CNN Layered Architecture

| Layer | Output Size | Param Number |
|---|---|---|
| Embedding | 1000 x 50 | 2157800 |
| Conv1D | 999 x 32 | 3232 |
| Dropout | 999 x 32 | 0 |
| Global Maxpool 1D | 32 | 0 |
| Dense | 2 | 66 |

## 3.3 BiLSTM

BiLSTM is a development of the LSTM modelwhere there are two layers that process each otherreverse direction, this model is very good to recognizepatterns in sentences as each word in the documentprocessed sequentially. The layer underneathmoves forward (forward), namely understanding and processingfrom the first word to the last word while the layerabove it moves backward (backward), namely understanding andprocessing from the last word to the first word. Withthere is a layer of two opposing directions thenthe model can understand and take the perspective of the wordpreceding and forefront, so the learning processwill be deeper which impact on the model will be moreunderstand the context in the document. Detail of The Bi-LSTM architecture that we use in this study can be seen in the Table 3.

Table3: Bi-LSTM Layered Architecture

| Layer | Output Size | Param Number |
|---|---|---|
| Embedding | 1000 x 50 | 2157800 |
| Bidirectional | 1000 x 256 | 184320 |
| Global Maxpool 1D | 256 | 0 |
| Dropout | 256 | 0 |
| Dense | 2 | 514 |

## 3.4 Hybrid CNN-BiLSTM

Another approach to deep learning is to combine models with each other. In this study we tried to combine the CNN and BiLSTM models. The details of the Hybrid CNN-BiLSTM architecture can be seen in the Table 4.

Table4: Hybrid CNN-Bi-LSTM Layered Architecture

| Layer | Output Size | Param Number |
| --- | --- | --- |
| Embedding | 1000 x 50 | 2157800 |
| Conv1D | 999 x 32 | 3232 |
| Dropout | 999 x 32 | 0 |
| Maxpool1D | 499 x 32 | 0 |
| Bidirectional | 499 x 256 | 165888 |
| Global Maxpool 1D | 256 | 0 |
| Dropout | 256 | 0 |
| Dense | 2 | 514 |

After discussing the entire deep learning architecture that will be used, we will then show the hyperparameters to be used for deep learning models which can be seen in the Table 5.

Table5: Hyperparameter for BERT

| Hyperparameter | Value |
| --- | --- |
| Loss function | Categorical-crossentropy |
| Learning rate | 2e-05 |
| Optimizer | Adam |
| Number of epochs | 50 |
| Batch size | 16 |
| Dropout | 0.5 |

## 4 EXPERIMENTAL SETUP

### 4.1 Dataset details

This research uses 3 datasets that are combined into one. The first dataset is a dataset obtained from Mendeley with 600 data, containing 372 valid news stories and 228 fake news. The dataset can be accessed on the https://data.mendeley.com/datasets/p3hfgr5j3m/1. The next dataset is a dataset derived from github, Hoax News Classification. The dataset that can be accessed at https://github.com/pierobeat/Hoax-News-Classification has 500 data with 250 valid news and 250 fake news. While the last dataset is a dataset obtained through scrapping on https://turnbackhoax.id page. Starting from the 25th April 2021, this page has 35 categories with a total of 6973 data. By considering the content of each category, we use the "Benar" category as valid news with 433 data and the "Hoax" category with 683 data as fake news. The scrapping results can be accessed at https://github.com/JibranFawaid/turnbackhoax-dataset. The features that we use from the three datasets are the Label, Headline, and Body of each news. There are examples of fake news and valid news that we have studied. An example of fake news is like this

"*Siapa yang tidak tahu dengan lele, ikan yang satu ini merupakan favorit bagi banyak orang karena memiliki rasa yang enak dan memiliki tekstur daging yang lembut namun ternyata lele ikan paling kotor…*" while an example of a valid news is like this

*"Mungkin Anda pernah membaca artikel atau sekadar pesan berantai soal pertolongan pertama penderita stroke dengan menusukkan jarum ke ujung jari hingga berdarah. Jangan menelan mentah informasi ini dan melakukannya. Diterangkan dr Astried Indrasari SpPD…"*

## 4.2 Preprocessing of Dataset

Using 3 datasets that have been combined previously, we divide the training data and testing data with a ratio of 80:20. As the result, the data is divided into 1772 on training data and 444 on testing data. The next stage, we perform a cleaning process on the dataset using the regex library. This method is functionally for correcting the data through an algorithm. So that it can save costs and time to be used next. As we know, that in the dataset, there are words that need to be ignored in processing, such as the words "dan", "atau", "tapi", etc. So, we do a Stopword Removal to delete these words to increase the processing speed and performance that will be carried out using the Sastrawi library.

## 4.3 Word Embedding

All neural network methods use word embedding with word2vec with 50 dimensions. Word embedding is used as a technique to convert a word into a vector or array consisting of a collection of numbers. The Gensim library [21] was chosen as the word embedding process. Gensim is a library that can be used for processing natural language and unsupervised topic modeling.

## 4.4 Transformers

The fine tuning strategy in BERT uses an existing pre-trained model. Pre-trained BERT-Multilingual-Cased of hugging face [19] was chosen in this study because it provides various languages including Bahasa. The details of the BERT-Multilingual-Cased layer can be seen in Table 6.

Table 6: BERT-Multilingual-Cased

| Details | Value |
| --- | --- |
| Language | 104 |
| Layer | 12 |
| Hidden Layer | 768 |
| Head | 12 |
| Parameter | 110 |

## 5 RESULTS AND DISCUSSION

In this study, we use three data sources with a total data reaching 2,216 with a distribution of 1,055 as valid data and 1,161 as false data. The dataset is then divided into data with a ratio of 80%: 20% for training data and test data sequentially. So that the total training data used is 1772 and 444 test data. In this study using Google Colab tools. This research uses the BERT model as the research baseline and is compared with deep learning algorithms such as CNN, BiLSTM, and Hybrid CNN-BiLSTM. Comparison of research can be seen in Table 7. From the results of the research that has been done, it is evident that in general the BERT architecture used produces a classifier that is better than the other classifiers.

In general, the difference between BERT performance and others is quite significant, for example on accuracy. BERT accuracy can reach 90% while CNN is only 74%. Where at least there is an increase in accuracy of up to 16%. Apart from accuracy, significant differences can also be seen in other metrics such as precision and f1-score. This proves that the BERT architecture is better in terms of natural language processing, especially for fake news detection.

Table 7: Comparison of Neural Network and Transformer

| Performance (%) | Deep Learning Model | | | |
| --- | --- | --- | --- | --- |
| | CNN | BiLSTM | Hybrid CNN-BiLSTM | BERT |
| Accuracy | 74% | 85% | 74% | **90%** |
| Macro Precsion | 74% | 85% | 78% | **90%** |
| Macro Recall | 74% | 85% | 73% | **90%** |
| Macro F1-Score | 74% | 85% | 72% | **90%** |
| Precision | 79% | 88% | 68% | **90%** |
| Recall | 69% | 82% | **94%** | 91% |
| F1-Score | 74% | 85% | 79% | **91%** |

## 6 CONCLUSION

Problems regarding the spread of fake news occur in various countries including Indonesia. This research proves that the problem of detecting fake news in Indonesian can be solved using a deep learning model. This research shows that the purposed method produces a better model when compared to other models. BERT as a "state-of-the-art" model is able to outperform other models in various performance metrics such as accuracy, precision, and F1-score.

**ACKNOWLEDGMENTS**

This work would not have been possible without the support of the Faculty of Computer Science, Brawijaya University